\pdfoutput=1

\documentclass[11pt]{article}

\usepackage{EMNLP2023}

\usepackage{times}
\usepackage{latexsym}
\usepackage{graphicx}

\usepackage[T1]{fontenc}

\usepackage[utf8]{inputenc}

\usepackage{microtype}

\usepackage{inconsolata}

\usepackage{enumitem}
\usepackage{pgfplots}
\usepackage{tabularx}
\usepackage{caption}
\usepackage{subcaption}
\usepackage{bbm}
\usepackage{url}
\usepackage{amsmath} 
\usepackage{pifont}
\usepackage{arydshln}
\usepackage{makecell}
\usepackage{booktabs}
\usepackage{multirow} 
\usepackage{mathtools}
\title{Rethinking the Generation of High-Quality CoT Data from the Perspective of LLM-Adaptive Question Difficulty Grading}

\author{%
Qianjin Yu$^{1}$ \quad Keyu Wu$^{1}$ \quad Zihan Chen$^1$ \quad Chushu Zhang$^1$ \quad ManLin Mei $^1$  \\
\textbf{Lingjun Huang}$^{1}$ \quad \textbf{Fang Tan}$^1$ \quad \textbf{Yongsheng Du}$^1$ \quad \textbf{Kunlin Liu}$^1$ \quad \textbf{Yurui Zhu}$^1$ \\
$^1$Intelligent System Department, Zhongxing Telecom Equipment(ZTE), Changsha, Hunan, China\\
\texttt{yuqianjin58@gmail.com, wukeyu9029@gmail.com}
}
\pgfplotsset{compat=1.18}
\begin{document}
\maketitle
\begin{abstract}
%
Recently, DeepSeek-R1 (671B)~\cite{deepseekai2025deepseekr1incentivizingreasoningcapability} has demonstrated its excellent reasoning ability in complex tasks and has publicly shared its methodology.
This provides potentially high-quality chain-of-thought (CoT) data for stimulating the reasoning abilities of small-sized large language models (LLMs).
To generate high-quality CoT data for different LLMs,
we seek an efficient method for generating high-quality CoT data with LLM-Adaptive question difficulty levels.
First, we grade the difficulty of the questions according to the reasoning ability of the LLMs themselves and construct an LLM-Adaptive question database. 
Second, we sample the problem database based on a distribution of difficulty levels of the questions and then use DeepSeek-R1 (671B)~\cite{deepseekai2025deepseekr1incentivizingreasoningcapability} to generate the corresponding high-quality CoT data with correct answers.
Thanks to the construction of CoT data with LLM-Adaptive difficulty levels, we have significantly reduced the cost of data generation and enhanced the efficiency of model supervised fine-tuning (SFT).
Finally, we have validated the effectiveness and generalizability of the proposed method in the fields of complex mathematical competitions and code generation tasks.
Notably, with only 2k high-quality mathematical CoT data, our ZMath-32B surpasses DeepSeek-Distill-32B in math reasoning task. Similarly, with only 2k high-quality code CoT data, our ZCode-32B surpasses DeepSeek-Distill-32B in code reasoning tasks.
\end{abstract}
\section{Introduction}
\begin{figure}[t]
\centering 
\includegraphics[width=1.0\linewidth]{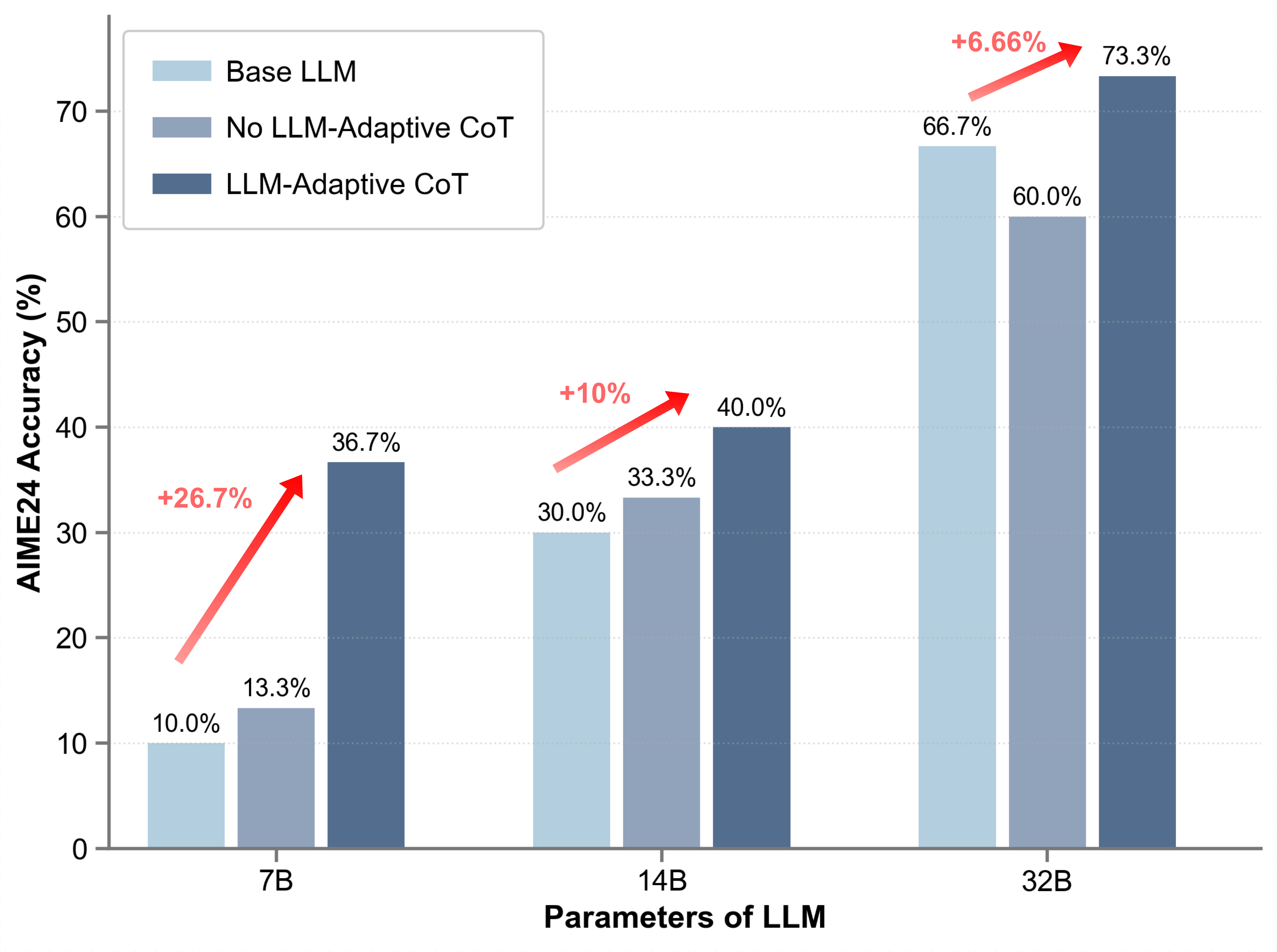}
\caption{Constrcution of CoT Data with/without LLM-Adaptive question difficulty grading. For LLms of different parameters, the former consistently outperforms the latter in reasoning performance on the mathematical competition dataset AIME24~\cite{aime}.}
\label{fig:introduction}
\vspace{-6mm}
\end{figure}
Since the release of DeepSeek-R1~\cite{deepseekai2025deepseekr1incentivizingreasoningcapability}, long chain-of-thought reasoning has gained widespread popularity in both foundational AI LLMs and a wide range of industrial AI applications.
However, the deployment of full-capacity R1-class models (e.g., DeepSeek-R1 with 671B parameters) poses substantial computational challenges, rendering their utilization infeasible for edge devices and real-time systems due to prohibitive resource demands.
This limitation has spurred intensive research into developing compact (<70B parameters) models capable of sustaining extended CoT reasoning, which is a core competency requirement for mathematical problem-solving, code generation, and scientific analysis.
Thanks to the shared reasoning process of DeepSeek-R1, we can get high-quality CoT data to boost the reasoning abilities of small-parameter LLMs.
Recently, many methods for generating CoT data based on DeepSeek-R1 have been widely studied in the community. ~\cite{bespoke_stratos,openthoughts} enhance the reasoning capabilities of LLMs by using massive CoT data, enabling their reasoning abilities to reach competitive levels. 
~\cite{ye2025limoreasoning,muennighoff2025s1simpletesttimescaling} aim to trigger the reasoning capabilities of large models by constructing a small batch of high-quality CoT data, yet they are unable to achieve further improvements in reasoning performance.
~\cite{wen2025light} focuses on refining reasoning abilities through multi-stage curriculum learning and rejection sampling. 
However, these approaches rarely consider the adaptive relationship between the Base LLM and its training data during data distillation.
Therefore, we rethink the question:\textbf{"What constitutes high-quality CoT data?"} and provide a comprehensive answer from the perspective of LLM-Adaptive Question Difficulty Grading.
Based on the above discussion, we propose a method for constructing high-quality CoT data based on LLM-Adaptive Question Difficulty Grading, as shown in Figure ~\ref{fig:introduction}.
Our method efficiently creates LLM-Adaptive CoT datasets, significantly enhancing reasoning abilities of LLMs across varying parameters without requiring resource-intensive fine-tuning approaches such as curriculum learning or rejection sampling. In contrast, LLMs trained on data without adaptive difficulty grading struggle to improve or may experience degraded performance under the same cost constraints. 
First, we evaluate and grade the difficulty levels of the reasoning questions by analyzing the intrinsic reasoning capabilities of the LLMs.
Based on this adaptive difficulty grading, we develop an adaptive question database that covers various difficulty levels.
Next, we sample questions from this adaptive library, guided by a carefully designed distribution across different levels of difficulty.
Finally, utilizing the powerful reasoning capabilities of the DeepSeek-R1 (671B)~\cite{deepseekai2025deepseekr1incentivizingreasoningcapability}, we generate corresponding high-quality CoT data that covers both mathematical reasoning and code generation tasks.
\noindent In summary, the main contributions of this work are as follows:
\begin{itemize}[leftmargin=*]
    \item \textbf{Adaptive Difficulty Evaluation:} We analyze the intrinsic reasoning capabilities of LLMs to effectively evaluate and classify reasoning questions into adaptive difficulty levels.
    
    \item \textbf{Comprehensive Adaptive Problem Library:} Based on the adaptive difficulty levels, we construct an extensive problem library covering diverse difficulty categories and carefully sample questions according to a well-designed difficulty distribution.
    
    \item \textbf{High-Quality CoT Data Generation:} Leveraging the DeepSeek-R1 model (671B)~\cite{deepseekai2025deepseekr1incentivizingreasoningcapability}, we generate high-quality chain-of-thought (CoT) datasets that cover mathematical reasoning and code generation tasks, ensuring consistent accuracy and detailed reasoning.

     \item \textbf{Comprehensive Evaluation:} We conduct extensive experiments on mathematical reasoning and code generation tasks using LLMs with different parameters, demonstrating the effectiveness and generalization of our proposed method for high-quality chain-of-thought (CoT) data generation.
\end{itemize}

\section{Related Work}
\subsection{Chain-of-Thought (CoT) Data Generation}
Current research focuses on three primary strategies for generating high-quality CoT data: (1) Manual annotation by domain experts to create gold-standard reasoning chains, primarily for benchmarking~\cite{li2024numinamath,huang2024olympicarena,gao2024omni}; (2) Prompt engineering leveraging LLMs' in-context learning capacity to elicit step-by-step rationales, though constrained by model biases~\cite{wu2024comparative,maiti2025comparative,whitney2024adaptive}; (3) Automated generation using self-alignment frameworks~\cite{mahene2024automated,liu2025llm}. While such methods show promise, particularly in boosting small sized LLMs via supervised fine-tuning, key challenges persist in ensuring the diversity, correctness, and coherence of generated reasoning chains~\cite{muennighoff2025s1simpletesttimescaling,ye2025limoreasoning}.
To address these limitations, recent advances integrate rejection sampling to filter low-quality reasoning paths and employ iterative refinement of teacher models. For instance, some approaches ~\cite{bespoke_stratos,openthoughts} leverage DeepSeek-R1 as the teacher reasoning model to improve step-by-step rationale generation, coupled with GPT-4o-mini for mathematical solution verification. Despite these improvements, scaling high-quality CoT generation across broader domains and difficulty levels remains an open challenge, particularly in maintaining robustness against error propagation in multi-step reasoning.
\subsection{LLM-Adaptive Difficulty Grading}
Traditional data generation approaches typically rely on static difficulty labels or heuristic rules, which inadequately account for the continuously evolving capabilities of large language models (LLMs). Inspired by adaptive assessment techniques in educational settings, this strategy automatically calibrates training data to align with the model’s current competence, thereby optimizing learning efficiency. Prior studies have explored alternative methods, such as employing LLM-generated scoring to adjust difficulty~\cite{sky_t1_2025,xie2024grade,lee2024college} or adopting curriculum learning frameworks~\cite{wen2025light,min2024imitate,yuan2025agent} that treat long-form QA as inherently challenging tasks. However, these approaches suffer from critical limitations, including inaccurate difficulty categorization and insufficient granularity in difficulty stratification. For instance, coarse-grained curriculum designs often oversimplify difficulty levels (e.g., categorizing questions merely by length), while LLM-based scoring methods struggle to capture nuanced reasoning demands. Such shortcomings highlight the need for more sophisticated, fine-grained adaptive frameworks to bridge the gap between data difficulty and model capability.

\section{Method}

\begin{figure*}[ht]
\centering 
\includegraphics[width=1.0\linewidth]{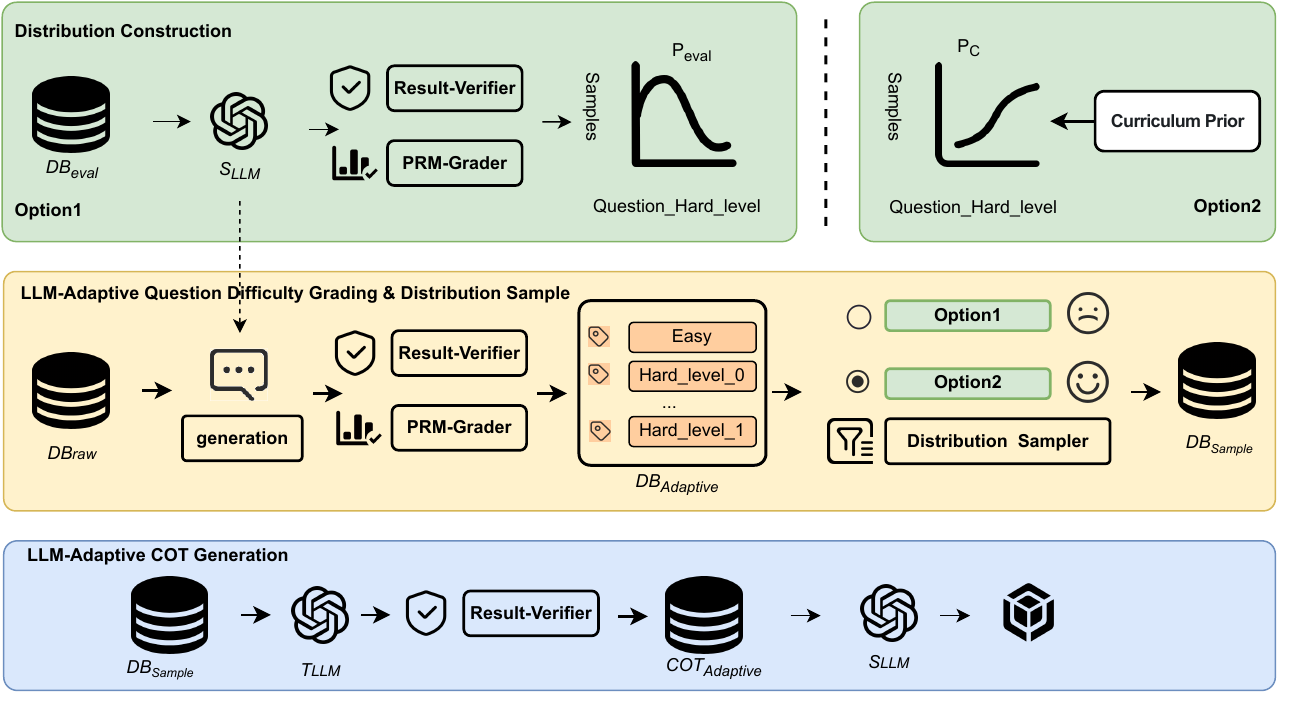}
\caption{The Framework for CoT Data Generation via LLM-Adaptive Question Difficulty Grading , comprising three core components: \textbf{Distribution Construction}, \textbf{LLM-Adaptive Question Difficulty Grading \& Distribution Sampling}, and \textbf{LLM-Adaptive Chain-of-Thought (CoT) Generation}
}
\label{fig:method}
\vspace{-5mm}
\end{figure*}

In this section, we introduce in detail our method for constructing high-quality CoT data with LLM-Adaptive question difficulty grading. As shown in ~\ref{fig:method}, our approach contains three components, described separately in the following subsections: (1) \textbf{Distribution Construction}, (2) \textbf{LLM-Adaptive Question Difficulty Grading \& Distribution Sampling}, and (3) \textbf{LLM-Adaptive CoT Generation}.
\subsection{Distribution Construction}
\label{sec:1}

To efficiently sample questions with model-adaptive difficulty grading, we require an effective reference distribution. To this end, we propose two alternative approaches for constructing a question-difficulty distribution. \textbf{Option1} leverages the Base LLM (defined as $S_{LLM}$) to characterize the true difficulty-level distribution (defined as $P_{eval}$) over the evaluation datasets.
\textbf{Option2} employs a customized distribution (defined as $P_{C}$) based on human-defined priors. Detailed descriptions of both methods are presented below.

\paragraph{Option1}
To obtain the actual difficulty-level distribution $P_{eval}$ from the $S_{LLM}$ for the evaluation data $DB_{eval}$, we first perform answer verification through a Result-Verifier. Those questions correctly answered by $S_{LLM}$ are defined as easy-level problems. Then, we utilize a PRM-Grader to grade the difficulty levels of the questions that the $S_{LLM}$ answers incorrectly. The specific grading formulation is shown as follows:  

\begin{equation}
P_{eval} =
\begin{cases}
\text{Easy}, & \text{if answer is \textup{True}} \\
\text{Grader}(Q, R), & \text{if answer is \textup{False}}
\end{cases}
\label{eq:distribution}
\end{equation}
where \text{Grader}(Q, R) denotes the Difficulty grading given by the PRM-Grader based on the reponse $R$ of the $S_{LLM}$. The details of the Result-Verifier and PRM-Grader can be seen in ~\ref{sec:2}

\paragraph{Option2}
Inspired by the idea of curriculum learning, we hypothesize that during the model fine-tuning process, the model learns relatively difficult questions more easily compared to very difficult ones. Therefore, we also propose a curriculum-learning-based customized distribution. Specifically, we classify question difficulty into five levels, with the number of samples at each difficulty level decreasing as the difficulty increases. The distribution can be formally defined by the following equation:
\begin{equation}
\begin{split}
P_{C} = \frac{N_i}{N_{\text{total}}} &=  \frac{w_i}{\sum_{j=1}^5 w_j}, \\
w_i &> w_{i+1},\quad i = 1, 2, \dots, 4
\end{split}
\label{eq:difficulty-sampling}
\end{equation}
where $N_{i}$ denotes the number of questions at difficulty level $i$, $N_{total}$ denotes the total number of questions, and $w_{i}$ represents the weight assigned to difficulty level $i$. The constraint $w_{i}>w_{i+1}$ illustrates that the assigned number of samples decreases as question difficulty increases.

We propose two methods for constructing model-adaptive difficulty distributions: Option1 derives difficulty levels from the actual performance of the $S_{LLM}$, while Option2 uses a curriculum-learning-inspired human-defined distribution. Subsequent experiments in section \ref{exp:1} will analyze and compare these approaches in detail. Next, we will use the distributions to guide the selection of LLM-Adaptive Questions.

\subsection{LLM-Adaptive Question Difficulty Grading \& Distribution Sampling}
\label{sec:2}
After constructing the question difficulty distribution based on either the evaluation-set distribution or curriculum-learning-inspired principles, 
we further need to build a specialized candidate question database from a large-scale data space using the model-adaptive difficulty grading method. 
Following such construction, we can perform sampling according to the predefined difficulty distribution to obtain the final high-quality questions.
First, we illustrate our LLM-Adaptive Question Difficulty Grading method for building the candidate question database (defined as $DB_{Adaptive}$) in detail.
Then, we describe the process of Distribution Sample to acquire the LLM-Adaptive Questions.
\paragraph{LLM-Adaptive Question Difficulty Grading}
First, we collect original questions from large-scale open-source datasets, each accompanied by standardized answers, thus constructing an initial question-answer database (defined as $DB_{raw}$). Next, we generate responses for these questions using the $S_{LLM}$ and record their reply trajectories and results.
Then, we apply the Result-Verifier customized to specific tasks. For mathematical reasoning, we adopt a Math Verifier, directly comparing the LLM-generated and standard answers; for code generation tasks, correctness is verified by executing the produced code against a suite of test cases, passing all tests indicating correctness.
According to equation ~\ref{eq:distribution} , verified correct responses are labeled as easy and directly added to the candidate question database. For responses deemed incorrect, we label these questions as difficult and further utilize the PRM-Grader, assigning difficulty levels into five categories. Specifically, the PRM-Grader computes an average score (ranging from 0 to 1) reflecting the response trajectory of $S_{LLM}$, mapping this score onto five discrete difficulty levels, with lower scores indicating relatively higher difficulty.
Ultimately, the questions categorized by these difficulty levels are collectively included within our candidate question database, thereby completing the construction of $DB_{Adaptive}$ in a LLM-Adaptive manner.
\paragraph{Distribution Sample}
After constructing the candidate question database, we employ a distribution-based sampler, guided by the question-difficulty distribution established in Section 3.1, to sample high-quality, model-adaptive questions from this database as preliminary inputs for obtaining high-quality CoT data. This procedure is formally defined as follows:
\begin{equation}
DB_{\text{sample}} \sim \text{Sampler}\left(DB_{\text{Adaptive}}, P_{\text{eval}} \lor P_{\text{C}} \right),
\end{equation}
where $DB_{\text{sample}}$ represents the sampled questions, $DB_{\text{Adaptive}}$ denotes the candidate question database, and $P_{\text{eval}}$, $P_{C}$ indicates the difficulty-level distribution determined in Section ~\ref{sec:1}.

\subsection{LLM-Adaptive CoT Generation}
\label{sec:3}
After obtaining the $DB_{Sample}$, we directly employ the $T_{LLM}$ to generate responses and associated reasoning processes for these sampled questions. Subsequently, we apply the Result-Verifier to examine and validate the correctness of these generated responses. The implementation of Result-Verifier here is identical to that described in Section ~\ref{sec:2}. The $T_{LLM}$ in our experiments is DeepSeek-R1(671B) ~\cite{deepseekai2025deepseekr1incentivizingreasoningcapability}.
Following the verification process, we select only those questions whose corresponding responses and reasoning processes have been validated as correct, thereby forming a high-quality CoT dataset $COT_{Adaptive}$. Finally, this rigorously-constructed CoT dataset serves as training data for fine-tuning $S_{LLM}$ to get the final reasoning LLM $R_{LLM}$.

\section{Experiments}
\label{exp:1}

\subsection{Setup}

\paragraph{Datasets and Metrics}
Our training datasets consist of high-quality mathematical reasoning problems sourced from NuminaMath~\cite{numina_math_datasets}, historical AIME problems~\cite{aime}, and OlympicArena~\cite{huang2024olympicarena}, as well as challenging code generation tasks from TACO~\cite{li2023taco} and CodeForces~\cite{penedo2025codeforces}.

\paragraph{Benchmarks}
To give a reasonable result, we evaluate our trained models
on the following authoritative benchmarks:
\begin{itemize}[leftmargin=0.5cm]
\item \emph{AIME24 and AIME25}~\cite{aime} comprise challenging mathematics competition problems from the American Invitational Mathematics Examination of 2024 and 2025. 
\item \emph{MATH500}~\cite{lightman2023lets} is a representative subset of 500 mathematical problems from the comprehensive MATH dataset. 
\item \emph{GPQA}~\cite{rein2024gpqa} is a dataset focused on graduate-level physics questions designed to evaluate advanced problem-solving skills. 
\item \emph{LiveCodeBench (EASY, MEDIUM, HARD)}~\cite{jain2024livecodebenchholisticcontaminationfree} contains competitive coding problems sourced from various platforms categorized by three difficulty levels.
\end{itemize}

\begin{table}
\small
\centering
\begin{tabular}{@{}cccccccc@{}}
\toprule
\multirow{2}{*}{Model} & \multicolumn{4}{c}{MATH}                                         \\ \cmidrule(l){2-5} 
 &
  \multicolumn{1}{l}{\begin{tabular}[c]{@{}l@{}}\emph{MATH}\\ \emph{500}\end{tabular}} &
  \multicolumn{1}{l}{\begin{tabular}[c]{@{}l@{}}\emph{AIME}\\ \emph{24}\end{tabular}} &
  \multicolumn{1}{l}{\begin{tabular}[c]{@{}l@{}}\emph{AIME}\\ \emph{25}\end{tabular}} &
  \multicolumn{1}{l}{\begin{tabular}[c]{@{}l@{}}\emph{GPQA}\end{tabular}} \\ \midrule
DS-distill-7B          & 89.4          & 56.67          & 33.3           & 49.49          \\
\textbf{Zmath-7B}               & \textbf{93.2} & \textbf{60}    & \textbf{43.33} & 49.49          \\ \midrule
phi4-14B               & 79.2          & 30             & 16.67          & 54.55          \\
\textbf{Zmath-14B}      & \textbf{89.4}          & \textbf{50.0}  & \textbf{36.67} & \textbf{63.13} \\ \midrule
DS-distill-32B         & 89.8          & 66.67          & 50.0           & 59.6           \\
Sky-32B-Preview        & 90   & 43.33          & 23.33          & 50.0           \\
\textbf{Zmath-32B}              & \textbf{94.6}          & \textbf{73.33} & \textbf{56.67} & \textbf{63.13} \\ \bottomrule
\end{tabular}
\caption{Comparison of LLMs with different parameters on Math Reasoning Benchmarks}
\label{tab:main_math}
\end{table}

\begin{table}
\small
\centering
\begin{tabular}{cccc}
\toprule
\multirow{2}{*}{Model} & \multicolumn{3}{c}{\emph{LiveCodeBench}}                     \\ \cmidrule(l){2-4} 
                       & \emph{EASY}           & \emph{MEDIUM}         & \emph{HARD}          \\ \midrule
DS-distill-7B     & 79.21           & \textbf{41.09}         & \textbf{11.11}          \\
\textbf{Zcode-7B}          & \textbf{81.0}           & 39.58          & 10.11 \\ \midrule
phi4-14B          & 72.4           & 29.91          & 5.19           \\
\textbf{Zcode-14B}         & \textbf{89.96} & \textbf{41.99}          & \textbf{8.89}  \\ \midrule
DS-distill-32B    & 92.11          & 74.92          & 30             \\
Sky-32B-Preview   & 84.23 & 46.53 & 8.89  \\
\textbf{Zcode-32B}      & \textbf{96.06} & \textbf{75.53} & \textbf{31.85} \\ \bottomrule
\end{tabular}
\caption{Comparison of LLMs with different parameters on Code Generation Benchmarks}
\label{tab:main_code}
\end{table}

\paragraph{Settings}
Our training framework builds on previous advancements in s1-1k\cite{muennighoff2025s1simpletesttimescaling}, LIMO\cite{ye2025limoreasoning}, and Light-R1\cite{lightr1proj}, implemented through the LLama-Factory\cite{zheng2024llamafactory} to leverage its proven scalability. The framework incorporates the Deepseek-R1 template, flash-attention2\cite{dao2023flashattention2} and Liger-Kernel\cite{hsu2024ligerkernelefficienttriton} to improve computational efficiency while minimizing memory requirements. All experiments are conducted on a 2×8 H800 GPU cluster, with performance evaluations executed using the Skythought benchmarking suite\cite{sky_t1_2025}. The core hyperparameters for the initial experiments included a context length of 16384, a learning rate of $\text{5e-6}$, a batch size of 128, and 10 training epochs.

\paragraph{Baselines}
We take the three representative baselines below for comparison:
\begin{itemize}[leftmargin=0.5cm]
\item \emph{phi-4}: phi-4\cite{abdin2024phi4} is a 14B LLM developed by Microsoft. Phi-4 demonstrates exceptional performance in complex reasoning tasks, particularly in mathematics, where it achieves 80.4\% on the MATH benchmark and 80.6\% on MGSM. 
\item \emph{DeepSeek-Distill-R1}: A series of LLMs developed by Deepseek, distilled from the Deepseek-R1  using 800k training instances\cite{deepseekai2025deepseekr1incentivizingreasoningcapability}. Our implementation primarily utilizes the 7B and 32B variants distilled on the Qwen architecture. 
\item \emph{Sky-T1-32B-Preview}: This model exhibits characteristics similar to Distill-R1-32B, but was developed by the Sky-T1 team using 10,000 distillation samples that specifically target mathematical and coding capabilities\cite{reduce_overthinking_2025}. 

\end{itemize}

\subsection{Results and Analysis}
\paragraph{Main Results}
Table \ref{tab:main_math} and Table \ref{tab:main_code} present the results of controlled experiments evaluating LLM performance on mathematical and coding benchmark datasets, respectively. Synthetic mathematical and coding data were generated using three base LLMs, DS-distill-7B, phi4-14B, and DS-distill-32B, and subjected to supervised fine tuning (SFT) to produce a series of LLMs at 7B, 14B, and 32B parameter scales. These LLMs were grouped into two categories: Zmath for mathematical tasks and Zcode for coding tasks.

The Zmath-14B LLM was trained on approximately 16,000 distilled data points derived from Deepseek-R1. The distilled data set was manually verified to ensure the accuracy of the answers. After training, the LLM output was aligned with the Deepseek-r1 format. Compared to the baseline phi4-14B LLM, Zmath-14B demonstrated substantial improvements: an average gain of 20 points on the AIME (2024, 2025) and Livecode-easy benchmarks; an approximate increase of 10 points on the GPQA and Livecode-medium datasets.Notably, Zmath-14B outperformed the Sky-32B-Preview LLM on mathematical tasks, highlighting its superior capability in this domain.

To explore the upper limit of our adaptive data synthesis approach, we trained LLMs using only 2,000 Chain of Thought (CoT) mathematical and coding data points on the DS-distill-32B. The results were as follows:
Zmath-32B achieved an average improvement of 5 points across all mathematical benchmarks, significantly exceeding the performance of DS-distill-32B.
On the Livecode Bench, Zcode-32B recorded an average gain of 2.14 points over DS-distill-32B. We hypothesize that further enhancements in coding performance, relative to mathematical gains, may necessitate a larger training corpus.
Comparable performance improvements were observed with the smaller Zmath-7B and Zcode-7B models, consistent with the trends observed in their 32B counterparts. These findings underscore the efficacy of our data synthesis method across varying model scales and task domains.
\paragraph{Ablation Studies}
We conducted four ablation studies using the code datasets to comprehensively validate the effectiveness of our proposed LLM-Adaptive difficulty grading approach from various perspectives.

\paragraph{(1) Comparison of Difficulty Grading Methods (PRM vs. UT)}
This experiment verifies the effectiveness of different difficulty grading methods. We compared two difficulty grading methods: Process Reward Model (PRM)-based grading, which assigns 0-1 scores divided into five levels (lower scores indicate higher difficulty), and Unit Test (UT)-based grading, which categorizes problems into five levels based on the percentage of passed test cases (lower pass rates indicate higher difficulty). As shown in Table \ref{tab:difficulty_grading}, we compare the results of DeepSeek-Distill-32B trained with UT-graded 2K data, PRM-graded 2K data, and the baseline evaluated on LiveCodeBench (easy-medium-hard), demonstrating the superior effectiveness of the PRM-based difficulty grading method.



\begin{table}
\small
\centering
\begin{tabular}{cccc}
\toprule
\multirow{2}{*}{Grading Methods} & \multicolumn{3}{c}{\emph{LiveCodeBench}}                     \\ \cmidrule(l){2-4} 
                       & \emph{EASY}           & \emph{MEDIUM}         & \emph{HARD}           \\ \midrule
No-Grading                            & 92.11          & 74.92          & 30.00             \\
UT-Grading                               & 93.19          & 71.60           & 28.15          \\
PRM-Grading                              & \textbf{96.06} & \textbf{75.53} & \textbf{31.85} \\ \bottomrule
\end{tabular}
\caption{Performance (\%) of DS-distill-32B trained on 2K data using different difficulty grading methods (None, UT-based, and PRM-based), evaluated on LiveCode-Bench.}
\label{tab:difficulty_grading}
\end{table}

\paragraph{(2) Distribution Transfer Experiment}
To examine whether different models exhibit unique preferences for difficulty distributions, we perform a distribution transfer experiment. Specifically, we transfer the PRM-based difficulty distribution derived from the DeepSeek-Distill-32B model to train the DeepSeek-Distill-7B model. We then compare its performance with a counterpart trained using the 7B model's own PRM-based distribution. As shown in Table~\ref{tab:distribution_transfer}, the 7B model benefits more from training on its self-derived difficulty distribution, suggesting that model-specific difficulty adaptation plays a critical role in performance optimization.

\begin{table}
\small
\centering
\begin{tabular}{@{}cccccccc@{}}
\toprule
\multirow{2}{*}{CoT Source} & \multicolumn{4}{c}{MATH}                                         \\ \cmidrule(l){2-5} 
 &
  \multicolumn{1}{l}{\begin{tabular}[c]{@{}l@{}}\emph{MATH}\\ \emph{500}\end{tabular}} &
  \multicolumn{1}{l}{\begin{tabular}[c]{@{}l@{}}\emph{AIME}\\ \emph{24}\end{tabular}} &
  \multicolumn{1}{l}{\begin{tabular}[c]{@{}l@{}}\emph{AIME}\\ \emph{25}\end{tabular}} &
  \multicolumn{1}{l}{\begin{tabular}[c]{@{}l@{}}\emph{GPQA}\end{tabular}} \\ \midrule
Baseline-7B & 89.40 & 56.67 & 33.30 & 49.49 \\
+CoT from 32B & 92.00 & 56.67 & 40.00 & 45.96 \\
+CoT from 7B & \textbf{93.20} & \textbf{60.00} & \textbf{43.33} & \textbf{49.49} \\
\bottomrule
\end{tabular}
\caption{Performance (\%) of 7B LLM with training data distributions derived from 32B and 7B LLMs on math benchmarks. The 7B/32B LLM is DS-distill-7B/32B.}
\label{tab:distribution_transfer}
\end{table}

\paragraph{(3) Influence of Training Data Size}
To investigate the influence of the size of the training data on reasoning performance, we trained the DeepSeek-Distill-32B model using 1K and 2K PRM-graded math examples. As shown in Table~\ref{tab:training_data_size}, increasing the training data size from 1K to 2K led to consistent performance improvements across all four math benchmarks.


\begin{table}
\small
\centering
\footnotesize  
\setlength\tabcolsep{4pt}  
\begin{tabular}{@{}cccccccc@{}}
\toprule
\multirow{2}{*}{Training Setup} & \multicolumn{4}{c}{MATH}                                         \\ \cmidrule(l){2-5} 
 &
  \multicolumn{1}{l}{\begin{tabular}[c]{@{}l@{}}\emph{MATH}\\ \emph{500}\end{tabular}} &
  \multicolumn{1}{l}{\begin{tabular}[c]{@{}l@{}}\emph{AIME}\\ \emph{24}\end{tabular}} &
  \multicolumn{1}{l}{\begin{tabular}[c]{@{}l@{}}\emph{AIME}\\ \emph{25}\end{tabular}} &
  \multicolumn{1}{l}{\begin{tabular}[c]{@{}l@{}}\emph{GPQA}\\\end{tabular}} \\ \midrule
No PRM fine-tuning & 89.80 & 66.67 & 50.00 & 59.60 \\
+1K PRM-graded data  & \textbf{95.50} & 73.33 & 53.33 & 60.61 \\
+2K PRM-graded data  & 94.60 & \textbf{73.33} & \textbf{56.67} & \textbf{63.13} \\
\bottomrule
\end{tabular}
\caption{Performance (\%) of DeepSeek-Distill-32B trained on varying sizes of PRM-graded math data.}
\label{tab:training_data_size}
\end{table}

\paragraph{(4) Impact of Training Data Distribution Standard}
We compare two strategies for defining training data distributions: \textbf{Option 1} leverages the base LLM to infer the true difficulty distribution from the evaluation dataset, while \textbf{Option 2} relies on human-defined prior distributions. As shown in Tab.~\ref{tab:distribution_standard}, using the model-inferred evaluation distribution (Option 1) achieves stronger results on most benchmarks, indicating that aligning training distribution with evaluation difficulty leads to better generalization.


\begin{table}[htbp]
\small
\centering
\footnotesize  
\setlength\tabcolsep{4pt}  
\begin{tabular}{@{}cccccccc@{}}
\toprule
\multirow{2}{*}{Sample Distribution} & \multicolumn{4}{c}{MATH}                                         \\ \cmidrule(l){2-5} 
 &
  \multicolumn{1}{l}{\begin{tabular}[c]{@{}l@{}}\emph{MATH}\\ \emph{500}\end{tabular}} &
  \multicolumn{1}{l}{\begin{tabular}[c]{@{}l@{}}\emph{AIME}\\ \emph{24}\end{tabular}} &
  \multicolumn{1}{l}{\begin{tabular}[c]{@{}l@{}}\emph{AIME}\\ \emph{25}\end{tabular}} &
  \multicolumn{1}{l}{\begin{tabular}[c]{@{}l@{}}\emph{GPQA}\\\end{tabular}} \\ \midrule
Baseline-7B & 89.40 & 56.67 & 33.30 & 49.49 \\
+Option 1 (LLM) & \textbf{93.20} &60.00 & \textbf{43.33} & \textbf{49.49} \\
+Option 2 (Human)  & 90.80 & \textbf{63.33} & 33.33 & 48.99 \\
\bottomrule
\end{tabular}
\caption{Performance (\%) of DS-distill-7B trained on 2K PRM-graded data using different training distribution strategies. Option 1 uses LLM-inferred evaluation-set distribution; Option 2 adopts human-defined priors.}
\label{tab:distribution_standard}
\end{table}
\section{Conclusion}
%

%
In this paper, we propose a general and efficient method for constructing high-quality Chain-of-Thought (CoT) datasets.
Firstly, we build a question database more aligned with the Base LLM itself by leveraging a method that adaptively grade question difficulty. This database possess a potential source of high-quality questions.
Next, we use the difficulty distribution from either LLM performance on evaluation datasets or curriculum-learning-inspired difficulty levels, to sample crucial questions for improving the reasoning capability.
Finally, these selected questions are employed to generate Chain-of-Thought data through the teacher LLM (DeepSeek-R1), forming a COT dataset that is adaptively graded according to question difficulty aligned with the Base LLM.
Benefiting from our constructed COT data, we effectively refine LLMs through supervised fine-tuning (SFT), achieving improved reasoning abilities across LLMs of different parameter scales.
In the future, we plan to integrate our approach for constructing high-quality COT data with reinforcement learning or reject sampling, further enhancing the reasoning abilities of the models.




\bibliography{custom}

\begin{thebibliography}{32}
\expandafter\ifx\csname natexlab\endcsname\relax\def\natexlab#1{#1}\fi

\bibitem[{Abdin et~al.(2024)Abdin, Aneja, Behl, Bubeck, Eldan, Gunasekar, Harrison, Hewett, Javaheripi, Kauffmann, Lee, Lee, Li, Liu, Mendes, Nguyen, Price, Rosa, Saarikivi, Salim, Shah, Wang, Ward, Wu, Yu, Zhang, and Zhang}]{abdin2024phi4}
Marah Abdin, Jyoti Aneja, Harkirat Behl, S{\'e}bastien Bubeck, Ronen Eldan, Suriya Gunasekar, Michael Harrison, Russell~J. Hewett, Mojan Javaheripi, Piero Kauffmann, James~R. Lee, Yin~Tat Lee, Yuanzhi Li, Weishung Liu, Caio C.~T. Mendes, Anh Nguyen, Eric Price, Gustavo~de Rosa, Olli Saarikivi, Adil Salim, Shital Shah, Xin Wang, Rachel Ward, Yue Wu, Dingli Yu, Cyril Zhang, and Yi~Zhang. 2024.
\newblock \href {https://arxiv.org/abs/2412.08905} {Phi-4 technical report}.
\newblock \emph{arXiv preprint arXiv:2412.08905}.

\bibitem[{Dao(2024)}]{dao2023flashattention2}
Tri Dao. 2024.
\newblock Flash{A}ttention-2: Faster attention with better parallelism and work partitioning.
\newblock In \emph{International Conference on Learning Representations (ICLR)}.

\bibitem[{DeepSeek-AI et~al.(2025)DeepSeek-AI, Guo, Yang, Zhang, Song, Zhang, Xu, Zhu, Ma, Wang, Bi, Zhang, Yu, Wu, Wu, Gou, Shao, Li, Gao, Liu, Xue, Wang, Wu, Feng, Lu, Zhao, Deng, Zhang, Ruan, Dai, Chen, Ji, Li, Lin, Dai, Luo, Hao, Chen, Li, Zhang, Bao, Xu, Wang, Ding, Xin, Gao, Qu, Li, Guo, Li, Wang, Chen, Yuan, Qiu, Li, Cai, Ni, Liang, Chen, Dong, Hu, Gao, Guan, Huang, Yu, Wang, Zhang, Zhao, Wang, Zhang, Xu, Xia, Zhang, Zhang, Tang, Li, Wang, Li, Tian, Huang, Zhang, Wang, Chen, Du, Ge, Zhang, Pan, Wang, Chen, Jin, Chen, Lu, Zhou, Chen, Ye, Wang, Yu, Zhou, Pan, Li, Zhou, Wu, Ye, Yun, Pei, Sun, Wang, Zeng, Zhao, Liu, Liang, Gao, Yu, Zhang, Xiao, An, Liu, Wang, Chen, Nie, Cheng, Liu, Xie, Liu, Yang, Li, Su, Lin, Li, Jin, Shen, Chen, Sun, Wang, Song, Zhou, Wang, Shan, Li, Wang, Wei, Zhang, Xu, Li, Zhao, Sun, Wang, Yu, Zhang, Shi, Xiong, He, Piao, Wang, Tan, Ma, Liu, Guo, Ou, Wang, Gong, Zou, He, Xiong, Luo, You, Liu, Zhou, Zhu, Xu, Huang, Li, Zheng, Zhu, Ma, Tang, Zha, Yan, Ren, Ren, Sha, Fu, Xu, Xie, Zhang,
  Hao, Ma, Yan, Wu, Gu, Zhu, Liu, Li, Xie, Song, Pan, Huang, Xu, Zhang, and Zhang}]{deepseekai2025deepseekr1incentivizingreasoningcapability}
DeepSeek-AI, Daya Guo, Dejian Yang, Haowei Zhang, Junxiao Song, Ruoyu Zhang, Runxin Xu, Qihao Zhu, Shirong Ma, Peiyi Wang, Xiao Bi, Xiaokang Zhang, Xingkai Yu, Yu~Wu, Z.~F. Wu, Zhibin Gou, Zhihong Shao, Zhuoshu Li, Ziyi Gao, Aixin Liu, Bing Xue, Bingxuan Wang, Bochao Wu, Bei Feng, Chengda Lu, Chenggang Zhao, Chengqi Deng, Chenyu Zhang, Chong Ruan, Damai Dai, Deli Chen, Dongjie Ji, Erhang Li, Fangyun Lin, Fucong Dai, Fuli Luo, Guangbo Hao, Guanting Chen, Guowei Li, H.~Zhang, Han Bao, Hanwei Xu, Haocheng Wang, Honghui Ding, Huajian Xin, Huazuo Gao, Hui Qu, Hui Li, Jianzhong Guo, Jiashi Li, Jiawei Wang, Jingchang Chen, Jingyang Yuan, Junjie Qiu, Junlong Li, J.~L. Cai, Jiaqi Ni, Jian Liang, Jin Chen, Kai Dong, Kai Hu, Kaige Gao, Kang Guan, Kexin Huang, Kuai Yu, Lean Wang, Lecong Zhang, Liang Zhao, Litong Wang, Liyue Zhang, Lei Xu, Leyi Xia, Mingchuan Zhang, Minghua Zhang, Minghui Tang, Meng Li, Miaojun Wang, Mingming Li, Ning Tian, Panpan Huang, Peng Zhang, Qiancheng Wang, Qinyu Chen, Qiushi Du, Ruiqi Ge, Ruisong
  Zhang, Ruizhe Pan, Runji Wang, R.~J. Chen, R.~L. Jin, Ruyi Chen, Shanghao Lu, Shangyan Zhou, Shanhuang Chen, Shengfeng Ye, Shiyu Wang, Shuiping Yu, Shunfeng Zhou, Shuting Pan, S.~S. Li, Shuang Zhou, Shaoqing Wu, Shengfeng Ye, Tao Yun, Tian Pei, Tianyu Sun, T.~Wang, Wangding Zeng, Wanjia Zhao, Wen Liu, Wenfeng Liang, Wenjun Gao, Wenqin Yu, Wentao Zhang, W.~L. Xiao, Wei An, Xiaodong Liu, Xiaohan Wang, Xiaokang Chen, Xiaotao Nie, Xin Cheng, Xin Liu, Xin Xie, Xingchao Liu, Xinyu Yang, Xinyuan Li, Xuecheng Su, Xuheng Lin, X.~Q. Li, Xiangyue Jin, Xiaojin Shen, Xiaosha Chen, Xiaowen Sun, Xiaoxiang Wang, Xinnan Song, Xinyi Zhou, Xianzu Wang, Xinxia Shan, Y.~K. Li, Y.~Q. Wang, Y.~X. Wei, Yang Zhang, Yanhong Xu, Yao Li, Yao Zhao, Yaofeng Sun, Yaohui Wang, Yi~Yu, Yichao Zhang, Yifan Shi, Yiliang Xiong, Ying He, Yishi Piao, Yisong Wang, Yixuan Tan, Yiyang Ma, Yiyuan Liu, Yongqiang Guo, Yuan Ou, Yuduan Wang, Yue Gong, Yuheng Zou, Yujia He, Yunfan Xiong, Yuxiang Luo, Yuxiang You, Yuxuan Liu, Yuyang Zhou, Y.~X. Zhu,
  Yanhong Xu, Yanping Huang, Yaohui Li, Yi~Zheng, Yuchen Zhu, Yunxian Ma, Ying Tang, Yukun Zha, Yuting Yan, Z.~Z. Ren, Zehui Ren, Zhangli Sha, Zhe Fu, Zhean Xu, Zhenda Xie, Zhengyan Zhang, Zhewen Hao, Zhicheng Ma, Zhigang Yan, Zhiyu Wu, Zihui Gu, Zijia Zhu, Zijun Liu, Zilin Li, Ziwei Xie, Ziyang Song, Zizheng Pan, Zhen Huang, Zhipeng Xu, Zhongyu Zhang, and Zhen Zhang. 2025.
\newblock \href {http://arxiv.org/abs/2501.12948} {Deepseek-r1: Incentivizing reasoning capability in llms via reinforcement learning}.

\bibitem[{Gao et~al.(2024)Gao, Song, Yang, Cai, Miao, Dong, Li, Ma, Chen, Xu et~al.}]{gao2024omni}
Bofei Gao, Feifan Song, Zhe Yang, Zefan Cai, Yibo Miao, Qingxiu Dong, Lei Li, Chenghao Ma, Liang Chen, Runxin Xu, et~al. 2024.
\newblock Omni-math: A universal olympiad level mathematic benchmark for large language models.
\newblock \emph{arXiv preprint arXiv:2410.07985}.

\bibitem[{Hsu et~al.(2024)Hsu, Dai, Kothapalli, Song, Tang, Zhu, Shimizu, Sahni, Ning, and Chen}]{hsu2024ligerkernelefficienttriton}
Pin-Lun Hsu, Yun Dai, Vignesh Kothapalli, Qingquan Song, Shao Tang, Siyu Zhu, Steven Shimizu, Shivam Sahni, Haowen Ning, and Yanning Chen. 2024.
\newblock \href {http://arxiv.org/abs/2410.10989} {Liger kernel: Efficient triton kernels for llm training}.
\newblock \emph{arXiv preprint arXiv:2410.10989}.

\bibitem[{Huang et~al.(2024)Huang, Wang, Xia, Li, Zou, Xu, Fan, Ye, Chern, Ye et~al.}]{huang2024olympicarena}
Zhen Huang, Zengzhi Wang, Shijie Xia, Xuefeng Li, Haoyang Zou, Ruijie Xu, Run-Ze Fan, Lyumanshan Ye, Ethan Chern, Yixin Ye, et~al. 2024.
\newblock Olympicarena: Benchmarking multi-discipline cognitive reasoning for superintelligent ai.
\newblock \emph{Advances in Neural Information Processing Systems}, 37:19209--19253.

\bibitem[{Jain et~al.(2024)Jain, Han, Gu, Li, Yan, Zhang, Wang, Solar-Lezama, Sen, and Stoica}]{jain2024livecodebenchholisticcontaminationfree}
Naman Jain, King Han, Alex Gu, Wen-Ding Li, Fanjia Yan, Tianjun Zhang, Sida Wang, Armando Solar-Lezama, Koushik Sen, and Ion Stoica. 2024.
\newblock \href {http://arxiv.org/abs/2403.07974} {Livecodebench: Holistic and contamination free evaluation of large language models for code}.

\bibitem[{Labs(2025)}]{bespoke_stratos}
Bespoke Labs. 2025.
\newblock Bespoke-stratos: The unreasonable effectiveness of reasoning distillation.
\newblock https://www.bespokelabs.ai/blog/bespoke-stratos-the-unreasonable-effectiveness-of-reasoning-distillation.
\newblock Accessed: 2025-01-22.

\bibitem[{Lee and Song(2024)}]{lee2024college}
Jung~X Lee and Yeong-Tae Song. 2024.
\newblock College exam grader using llm ai models.
\newblock In \emph{2024 IEEE/ACIS 27th International Conference on Software Engineering, Artificial Intelligence, Networking and Parallel/Distributed Computing (SNPD)}, pages 282--289. IEEE.

\bibitem[{Li et~al.(2024)Li, Beeching, Tunstall, Lipkin, Soletskyi, Huang, Rasul, Yu, Jiang, Shen et~al.}]{li2024numinamath}
Jia Li, Edward Beeching, Lewis Tunstall, Ben Lipkin, Roman Soletskyi, Shengyi Huang, Kashif Rasul, Longhui Yu, Albert~Q Jiang, Ziju Shen, et~al. 2024.
\newblock Numinamath: The largest public dataset in ai4maths with 860k pairs of competition math problems and solutions.
\newblock \emph{Hugging Face repository}, 13:9.

\bibitem[{LI et~al.(2024)LI, Beeching, Tunstall, Lipkin, Soletskyi, Huang, Rasul, Yu, Jiang, Shen, Qin, Dong, Zhou, Fleureau, Lample, and Polu}]{numina_math_datasets}
Jia LI, Edward Beeching, Lewis Tunstall, Ben Lipkin, Roman Soletskyi, Shengyi~Costa Huang, Kashif Rasul, Longhui Yu, Albert Jiang, Ziju Shen, Zihan Qin, Bin Dong, Li~Zhou, Yann Fleureau, Guillaume Lample, and Stanislas Polu. 2024.
\newblock Numinamath.
\newblock \url{[https://huggingface.co/AI-MO/NuminaMath-CoT](https://github.com/project-numina/aimo-progress-prize/blob/main/report/numina_dataset.pdf)}.

\bibitem[{Li et~al.(2023)Li, Fu, Zhang, Huang, Sun, Lyu, Liu, Jin, and Li}]{li2023taco}
Rongao Li, Jie Fu, Bo-Wen Zhang, Tao Huang, Zhihong Sun, Chen Lyu, Guang Liu, Zhi Jin, and Ge~Li. 2023.
\newblock Taco: Topics in algorithmic code generation dataset.
\newblock \emph{arXiv preprint arXiv:2312.14852}.

\bibitem[{Lightman et~al.(2023)Lightman, Hunter, Kosaraju, Burda, Edwards, Baker, Lee, Leike, Schulman, Sutskever, and Cobbe}]{lightman2023lets}
Lightman, Hunter, Vineet Kosaraju, Yura Burda, Harri Edwards, Bowen Baker, Teddy Lee, Jan Leike, John Schulman, Ilya Sutskever, and Karl Cobbe. 2023.
\newblock Let's verify step by step.
\newblock \emph{arXiv preprint arXiv:2305.20050}.

\bibitem[{Liu et~al.(2025)Liu, Cui, Hu, Li, Lin, and Zhang}]{liu2025llm}
Beiming Liu, Zhizhuo Cui, Siteng Hu, Xiaohua Li, Haifeng Lin, and Zhengxin Zhang. 2025.
\newblock Llm evaluation based on aerospace manufacturing expertise: Automated generation and multi-model question answering.
\newblock \emph{arXiv preprint arXiv:2501.17183}.

\bibitem[{Mahene et~al.(2024)Mahene, Pereira, Kowalski, Novak, Moretti, and Laurent}]{mahene2024automated}
Anthony Mahene, Daniel Pereira, Vincent Kowalski, Elizabeth Novak, Catherine Moretti, and Josephine Laurent. 2024.
\newblock Automated dynamic data generation for safety alignment in large language models.
\newblock \emph{Authorea Preprints}.

\bibitem[{Maiti et~al.(2025)Maiti, Adewumi, Tikure, Wang, Sengupta, Sukhanova, and Jana}]{maiti2025comparative}
Aniruddha Maiti, Samuel Adewumi, Temesgen~Alemayehu Tikure, Zichun Wang, Niladri Sengupta, Anastasiia Sukhanova, and Ananya Jana. 2025.
\newblock Comparative analysis of openai gpt-4o and deepseek r1 for scientific text categorization using prompt engineering.
\newblock \emph{arXiv preprint arXiv:2503.02032}.

\bibitem[{Min et~al.(2024)Min, Chen, Jiang, Chen, Deng, Hu, Tang, Wang, Cheng, Song et~al.}]{min2024imitate}
Yingqian Min, Zhipeng Chen, Jinhao Jiang, Jie Chen, Jia Deng, Yiwen Hu, Yiru Tang, Jiapeng Wang, Xiaoxue Cheng, Huatong Song, et~al. 2024.
\newblock Imitate, explore, and self-improve: A reproduction report on slow-thinking reasoning systems.
\newblock \emph{arXiv preprint arXiv:2412.09413}.

\bibitem[{Muennighoff et~al.(2025)Muennighoff, Yang, Shi, Li, Fei-Fei, Hajishirzi, Zettlemoyer, Liang, Candès, and Hashimoto}]{muennighoff2025s1simpletesttimescaling}
Niklas Muennighoff, Zitong Yang, Weijia Shi, Xiang~Lisa Li, Li~Fei-Fei, Hannaneh Hajishirzi, Luke Zettlemoyer, Percy Liang, Emmanuel Candès, and Tatsunori Hashimoto. 2025.
\newblock \href {http://arxiv.org/abs/2501.19393} {s1: Simple test-time scaling}.

\bibitem[{of~America(2024)}]{aime}
Mathematical~Association of~America. 2024.
\newblock \href {https://artofproblemsolving.com/wiki/index.php/AIME_Problems_and_Solutions/} {Aime}.

\bibitem[{Penedo et~al.(2025)Penedo, Lozhkov, Kydlíček, Allal, Beeching, Lajarín, Gallouédec, Habib, Tunstall, and von Werra}]{penedo2025codeforces}
Guilherme Penedo, Anton Lozhkov, Hynek Kydlíček, Loubna~Ben Allal, Edward Beeching, Agustín~Piqueres Lajarín, Quentin Gallouédec, Nathan Habib, Lewis Tunstall, and Leandro von Werra. 2025.
\newblock Codeforces.
\newblock \url{https://huggingface.co/datasets/open-r1/codeforces}.

\bibitem[{Rein et~al.(2024)Rein, Hou, Stickland, Petty, Pang, Dirani, Michael, and Bowman}]{rein2024gpqa}
David Rein, Betty~Li Hou, Asa~Cooper Stickland, Jackson Petty, Richard~Yuanzhe Pang, Julien Dirani, Julian Michael, and Samuel~R. Bowman. 2024.
\newblock \href {https://openreview.net/forum?id=Ti67584b98} {{GPQA}: A graduate-level google-proof q\&a benchmark}.
\newblock In \emph{First Conference on Language Modeling}.

\bibitem[{Team(2025{\natexlab{a}})}]{sky_t1_2025}
NovaSky Team. 2025{\natexlab{a}}.
\newblock Sky-t1: Train your own o1 preview model within 450.
\newblock https://novasky-ai.github.io/posts/sky-t1.
\newblock Accessed: 2025-01-09.

\bibitem[{Team(2025{\natexlab{b}})}]{reduce_overthinking_2025}
NovaSky Team. 2025{\natexlab{b}}.
\newblock Think less, achieve more: Cut reasoning costs by 50
\newblock https://novasky-ai.github.io/posts/reduce-overthinking.
\newblock Accessed: 2025-01-23.

\bibitem[{Team(2025{\natexlab{c}})}]{openthoughts}
OpenThoughts Team. 2025{\natexlab{c}}.
\newblock {Open Thoughts}.
\newblock https://open-thoughts.ai.

\bibitem[{Wen et~al.(2025{\natexlab{a}})Wen, Cai, Xiao, He, An, Duan, Du, Liu, Tang, Lv, Zou, Deng, Jia, and Zhang}]{wen2025light}
Liang Wen, Yunke Cai, Fenrui Xiao, Xin He, Qi~An, Zhenyu Duan, Yimin Du, Junchen Liu, Lifu Tang, Xiaowei Lv, Haosheng Zou, Yongchao Deng, Shousheng Jia, and Xiangzheng Zhang. 2025{\natexlab{a}}.
\newblock Light-r1: Curriculum sft, dpo and rl for long cot from scratch and beyond.
\newblock \emph{arXiv preprint arXiv:2503.10460}.

\bibitem[{Wen et~al.(2025{\natexlab{b}})Wen, Cai, Xiao, He, An, Duan, Du, Liu, Tang, Lv, Zou, Deng, Jia, and Zhang}]{lightr1proj}
Liang Wen, Yunke Cai, Fenrui Xiao, Xin He, Qi~An, Zhenyu Duan, Yimin Du, Junchen Liu, Lifu Tang, Xiaowei Lv, Haosheng Zou, Yongchao Deng, Shousheng Jia, and Xiangzheng Zhang. 2025{\natexlab{b}}.
\newblock Light-r1: Curriculum sft, dpo and rl for long cot from scratch and beyond.
\newblock \emph{arXiv preprint arXiv:2503.10460}.

\bibitem[{Whitney et~al.(2024)Whitney, Jansen, Laskowski, and Barbieri}]{whitney2024adaptive}
Claire Whitney, Edward Jansen, Victor Laskowski, and Charles Barbieri. 2024.
\newblock Adaptive prompt regeneration and dynamic response structuring in large language models using the dynamic query-response calibration protocol.
\newblock \emph{Authorea Preprints}.

\bibitem[{Wu et~al.(2024)Wu, Peng, Du, Zheng, Liu, Wu, Ma, Li, Yang, Zhou et~al.}]{wu2024comparative}
Siwei Wu, Zhongyuan Peng, Xinrun Du, Tuney Zheng, Minghao Liu, Jialong Wu, Jiachen Ma, Yizhi Li, Jian Yang, Wangchunshu Zhou, et~al. 2024.
\newblock A comparative study on reasoning patterns of openai's o1 model.
\newblock \emph{arXiv preprint arXiv:2410.13639}.

\bibitem[{Xie et~al.(2024)Xie, Niu, Xue, and Guan}]{xie2024grade}
Wenjing Xie, Juxin Niu, Chun~Jason Xue, and Nan Guan. 2024.
\newblock Grade like a human: Rethinking automated assessment with large language models.
\newblock \emph{arXiv preprint arXiv:2405.19694}.

\bibitem[{Ye et~al.(2025)Ye, Huang, Xiao, Chern, Xia, and Liu}]{ye2025limoreasoning}
Yixin Ye, Zhen Huang, Yang Xiao, Ethan Chern, Shijie Xia, and Pengfei Liu. 2025.
\newblock \href {http://arxiv.org/abs/2502.03387} {Limo: Less is more for reasoning}.

\bibitem[{Yuan et~al.(2025)Yuan, Chen, Xi, Ye, Du, and Chen}]{yuan2025agent}
Siyu Yuan, Zehui Chen, Zhiheng Xi, Junjie Ye, Zhengyin Du, and Jiecao Chen. 2025.
\newblock Agent-r: Training language model agents to reflect via iterative self-training.
\newblock \emph{arXiv preprint arXiv:2501.11425}.

\bibitem[{Zheng et~al.(2024)Zheng, Zhang, Zhang, Ye, Luo, Feng, and Ma}]{zheng2024llamafactory}
Yaowei Zheng, Richong Zhang, Junhao Zhang, Yanhan Ye, Zheyan Luo, Zhangchi Feng, and Yongqiang Ma. 2024.
\newblock \href {http://arxiv.org/abs/2403.13372} {Llamafactory: Unified efficient fine-tuning of 100+ language models}.
\newblock In \emph{Proceedings of the 62nd Annual Meeting of the Association for Computational Linguistics (Volume 3: System Demonstrations)}, Bangkok, Thailand. Association for Computational Linguistics.

\end{thebibliography}
\bibliographystyle{acl_natbib}

\appendix



\end{document}